\def\year{2017}\relax
\documentclass[letterpaper]{article}
\usepackage{aaai17}
\usepackage{times}
\usepackage{helvet}
\usepackage{courier}
\usepackage{comment} %lan add
\usepackage{graphicx} %lan add
\usepackage{epstopdf} %lan add
\usepackage{amsmath} %zhu add
\usepackage{mathrsfs} %zhu add
\usepackage{scrextend} %zhu add
\usepackage{url} %lan add
\usepackage{subcaption} %lan add
\usepackage{epstopdf}%lan add
\usepackage{bm} %lan add
\usepackage{amssymb} %lan add
\usepackage{color}
\usepackage{flushend} %lan add
\usepackage{algorithm} %format of the algorithm
\usepackage{algorithmic} %format of the algorithm
\usepackage{multirow} %lan
\newcommand{\mathsfsl}{}
\begin{document}
\title{An End-to-End Spatio-Temporal Attention Model for  Human Action Recognition from Skeleton Data}

\author{
	{Sijie Song{\small $~^{1}$}, Cuiling Lan{\small $~^{2}$}\thanks{Corresponding author. This work was done at Microsoft Research Asia. This work was supported by National Natural Science Foundation of China under contract No. 61472011 and No. 61303178.}, Junliang Xing{\small $~^{3}$}, Wenjun Zeng{\small $~^{2}$}}, Jiaying Liu{\small $~^{1}$\large$^{*}$} \\
	$^{1}$\	Institute of Computer Science and Technology, Peking University, Beijing, China \\
    $^{2}$\,Microsoft Research Asia, Beijing, China \\
	$^{3}$\,Institute of Automation, Chinese Academy of Sciences, Beijing, China \\	
	\{ssj940920, liujiaying\}@pku.edu.cn,
	\{culan,wezeng\}@microsoft.com,
	jlxing@nlpr.ia.ac.cn
}

\maketitle
\begin{abstract}
Human action recognition is an important task in computer vision. Extracting discriminative spatial and temporal features to model the spatial and temporal evolutions of different actions plays a key role in accomplishing this task. In this work, we propose an end-to-end spatial and temporal attention model for human action recognition from skeleton data. We build our model on top of the Recurrent Neural Networks (RNNs) with Long Short-Term Memory (LSTM), which learns to selectively focus on discriminative joints of skeleton within each frame of the inputs and pays different levels of attention to the outputs of different frames. Furthermore, to ensure effective training of the network, we propose a regularized cross-entropy loss to drive the model learning process and develop a joint training strategy accordingly. Experimental results demonstrate the effectiveness of the proposed model, both on the small human action recognition dataset of SBU and the currently largest NTU dataset.
%We design joint-selection gates to explore the relative importance of joints within each frame and frame-selection gate to explore the importance of frames.

%spatio-temporal attention
%With the soft joint attention gates which explores the relative importances of joints, regularization item is designed to encourage the exploration of all joints and avoid morbidly trapping to local minimum. Finally, we propose an alternate training strategy for achieving the efficient combination of spatial and temporal attention mechanism. Experimental results consistently demonstrate the effectiveness of the proposed model, both on small human action recognition datasets and the currently largest NTU dataset.
%We build our spatio-temporal attention model on top of the Recurrent Neural Networks (RNNs) with Long Short-Term Memory, which is capable of modeling long-term temporal dependencies and learning feature representations automatically. On one hand, our spatial attention learning module drives the model to selectively focus on discriminative joints of skeleton within each frame of the inputs. On the other hand, our temporal attention learning module further drives the model to pay different attentions to the outputs of different frames.
\end{abstract}

\section{1~~Introduction}
\label{sec:introduction}
Recognition of human action is a fundamental yet challenging task in computer vision. It facilitates many applications such as intelligent video surveillance, human-computer interaction, video summary and understanding \cite{IVC10SurveyAction,CVIU11SurveyAction}. The key to the success of this task is how to extract discriminative spatial temporal features to effectively model the spatial and temporal evolutions of different actions.
%One crucial problem in this task is how to effectively model the spatial and temporal evolutions of different actions.
%One of the main challenges in this problem comes from the modeling of the spatial temporal evolutions of different actions.

One general approach  focuses on the recognition from RGB videos \cite{CVIU11SurveyAction}. Since each frame is a capture of the highly articulated human in a two-dimensional space, it loses some information of the three-dimensional (3D) space and then loses the flexibility of achieving human location and scale invariance. The other general approach leverages the high level information of skeleton data, which represents a person by the 3D coordinate positions of key joints (i.e., head, neck,$\cdots$, foot). Such representation is robust to variations of locations and viewpoints. Without combining RGB information, there is a lack of appearance information. Fortunately, biological observations from the early seminal work of Johansson suggest that the positions of a small number of joints can effectively represent human behavior even without appearance information \cite{PP73Perception}. Skeleton-based human representation has attracted increasing attention for recognizing human actions thanks to its high level representation and robustness to variations of locations and appearances \cite{han2016space}. The prevalence of cost-effective depth cameras such as Microsoft Kinect \cite{zhang2012microsoft} and the advance of a powerful human pose estimation technique from depth \cite{Shotton2011} make 3D skeleton data easily accessible. This boosts  research on skeleton-based human action recognition. In this work, we focus on recognition from skeleton data.
\begin{figure}[t]
	\centering
	\includegraphics[width=1\linewidth]{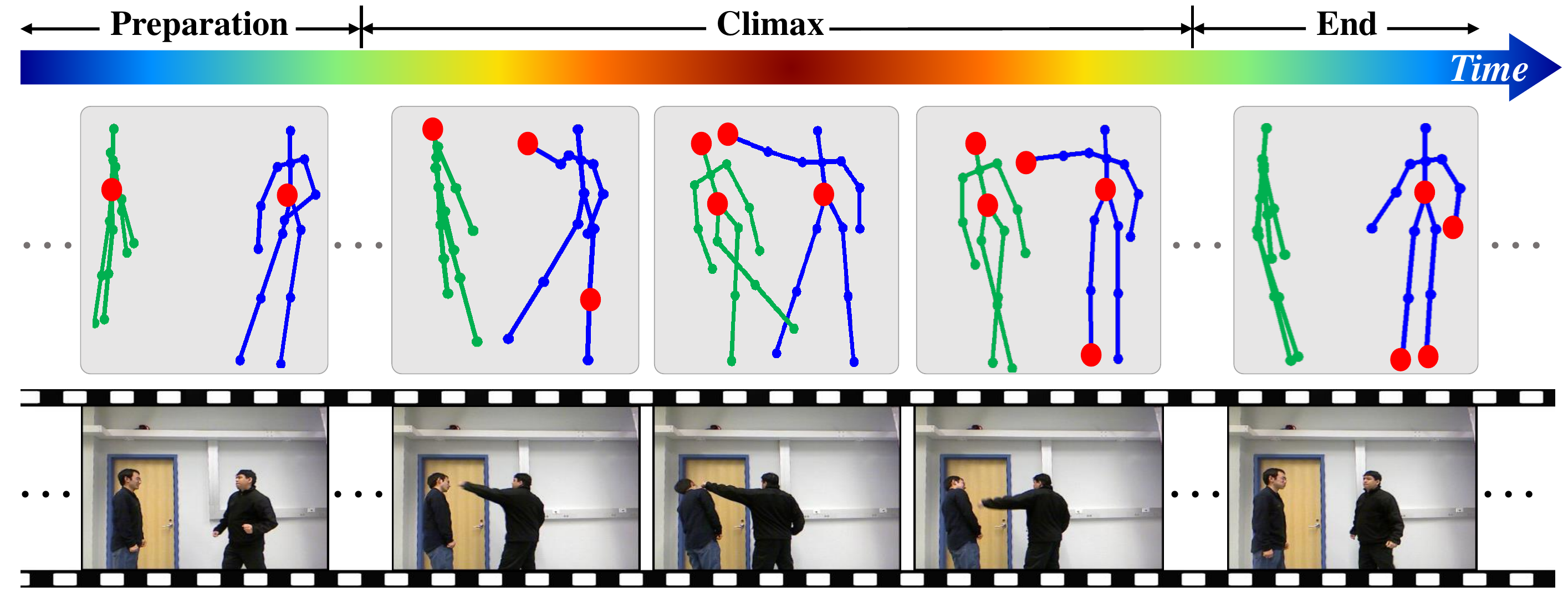}
\vspace{-5mm}
	\caption{Illustration of the procedure for an action ``punching". An action may experience different stages, and involve different discriminative subsets of joints (as the red circles). }
	%Best viewed in the color version.
	%\caption{Example of RGB and skeleton frames of \emph{Punching}. The flow of an action may experience different stages, and involve different subsets of joints (green). Best viewed in the color version.}	
\label{fig:skeleton}
\end{figure}

Fig. 1 shows an example of a series of skeleton frames (and RGB images) for the action ``punching". Each human body is represented by key joints in terms of coordinate positions in the 3D space. The articulated configurations of joints constitute various postures and a series of postures in a certain time order identifies an action. With the skeleton as an explicit high level representation of human pose, many works design algorithms taking the positions of joints as inputs. There are two basic components in these works. One is the design and mining of discriminative features from the skeleton, such as histograms of 3D joint locations (HOJ3D) \cite{CVPR12HO3DJ}, pairwise relative position features \cite{CVPR12Actionlet}, relative 3D geometry features \cite{vemulapalli2016r3dg}. The other is the modeling of temporal dynamics, such as Hidden Markov Model \cite{CVPR12HO3DJ}, Conditional Random Fields \cite{ICCV05CRF}, and Recurrent Neural Networks \cite{CVPR15HRNN}. In this work, we present a spatio-temporal attention model to incorporate the two components into an end-to-end deep learning architecture.
%relational pose features \cite{ToG05MotionRetrieval}
%For recognizing human actions, how to effectively explore and exploit the spatial and temporal evolution of video sequence is one of the key questions. To this end, we propose a spatial and temporal attention mechanism.

For spatial joints of skeleton, we propose a spatial attention module which conducts automatic mining of discriminative joints. A certain type of action is usually only associated with and characterized by the combinations of a subset of kinematic joints \cite{CVPR12Actionlet}. As the action proceeds, the associated joints may also change accordingly. For example, the joints ``hand", ``elbow", and ``head" are discriminative for the action ``drinking" while the joints from legs can be considered as noise. For an action ``approaching and shaking hands", at the beginning, the legs may be paid attention to; at the middle stage, the arms attract more attention. In contrast to actionlet \cite{CVPR12Actionlet}, the attentions to joints are allowed to vary over time, being content-dependent.
%the proposed spatial attention model adopts a spatial attention mechanism to conduct automatic mining of discriminative joints.

Furthermore, for a sequence of frames, we propose a temporal attention module which explicitly learns and allocates the content-dependent attentions to the output of each frame to boost recognition performance. For a sequence of some action, the flow of the action may experience different stages, e.g., the preparation, climax, and the end (Fig. \ref{fig:skeleton}). Taking the action ``punching" as an example, the two persons approach each other, stretch out the hands, and kick out the legs. The frames for identifying stretching out the hands and kicking out the legs are a part of the key sub-stage. Different sub-stages have different degrees of importance and robustness to variations. In this paper, in contrast to the ideas of extracting key frames \cite{carlsson2001action,BMVC08Information}, our proposed scheme pays different attentions to different frames instead of simply skipping frames.
%the proposed temporal attention model explicitly learns and allocates the content-dependent attentions over each frame to boost the recognition performance.
%The determination of the attention weights depends on both the current frame and the past frames.

%In other words, not all frames are equally important and different temporal frames may deserve different attentions in determining the action from a video segment.
%Taking the action of ``opening microwave" as an example, one may approach the microwave, stretch out one hand, pull back the hand. The frames for identifying pulling back the hand is part of the key sub-stage.

In summary, we have made the following four main contributions in this work.
%we have made four main contributions. % in this work
\begin{itemize}
\setlength{\itemsep}{0pt}\item An end-to-end framework with two types of attention modules is designed based on the LSTM networks for skeleton based human action recognition.

\setlength{\parsep}{0pt}\item A spatial attention module with joint-selection gates is designed to adaptively allocate different attentions to different joints of the input skeleton within each frame. A temporal attention module with frame-selection gate is designed to allocate different attentions to different frames.

\setlength{\parskip}{0pt}\item Spatio-temporal regularizations are proposed to enable the better learning of the networks.
%encourage the exploration of all joints.

\setlength{\parskip}{0pt}\item A joint training strategy is designed to efficiently train the entire end-to-end network.
%combine the spatial attention module and temporal attention module together.
% to the entire network
%We propose a joint training strategy to efficiently combine the spatial attention module and temporal attention module together to the entire network.
\end{itemize}

%Given a test sequence, our full model automatically determines which joints and which frames are more discriminative and assigns higher importance to them.
%we propose an end-to-end spatial and temporal attention based neural network for the task of human action recognition from skeleton data. We use multi-layered Recurrent Neural Networks (RNNs) with Long Short-Term Memory (LSTM) units for learning feature representations and modeling long-term temporal dependencies automatically.
%The determination of attention weights considers both the current frame contents and the memorized history information.
%, depending on the high level features of the current frame and the memorized history information

\section{2~~Related Work}
\label{sec:relatedwork}

\subsection{2.1~~Spatial Co-Occurrence Exploration}
\label{subsec:Cooccurrence}
An action is usually associated with and characterized by the interactions and combinations of a subset of skeleton joints. An actionlet ensemble model is proposed to mine such discriminative joints \cite{CVPR12Actionlet}, where an actionlet is a particular conjunction of the features for a subset of the joints and an action is represented as a linear combination of the actionlets. For example, for the action ``drinking", the subset of joints including ``hand", ``elbow", and ``head" composes a actionlet. Orderlet makes an extension of actionlet by including the feature of pairwise joint distance and allowing various sizes of a subset \cite{yu2014discriminative}. Actionlets or orderlets are mined from training samples for robust performance. In a recurrent neural network, a group sparsity constraint is introduced to the connection matrix to encourage the network to explore the co-occurrence of joints \cite{zhu2015co}.
% in the form of using different weights
%the idea of
% rather than a fixed size

In the above methods, once the mining is done, the degrees of importance of joints/features are fixed and there will be no change for different temporal frames and sequences. In contrast, our spatial attention module determines the degrees of importance of joints on the fly based on the contents.

\subsection{2.2~~Temporal Key Frame Exploration}
For identifying an action, not all frames in a sequence have the same importance. Some frames capture less meaningful information, or even carry misleading information associated with other types of actions, while some other frames carry more discriminative information \cite{liu2013boosted}. A number of approaches have proposed  using key frames as a representation for action recognition. One is to utilize the conditional entropy of visual words to measure the discriminative power of a given frame and the classification results from the top 25\% most discriminative frames are employed to make a majority vote for recognition \cite{BMVC08Information}. Another one employs the AdaBoost algorithm to select the most discriminative key frames for human action recognition \cite{liu2013boosted}. The learning of key frames can also be cast in a max-margin discriminative framework by treating them as latent variables \cite{raptis2013poselet}.
%Conditional entropy of the visual words is utilized to measure the discriminative power of a given frame  and the top 25\% most discriminative frames are selected \cite{BMVC08Information}. Then the majority vote from the classification results of key frames is employed for recognition. An AdaBoost learning algorithm is utilized to select the most discriminative key frames for human action recognition  \cite{liu2013boosted}. Raptis et al. cast the learning of key frames in a max-margin discriminative framework, by treating them as latent variable \cite{raptis2013poselet}.
%even describe a pose common to all action sequences

Leveraging key frames can help exclude noise frames, e.g., frames which are less relevant to the underlying actions. However, in comparisons to the holistic based approaches \cite {simonyan2014two,wu2015modeling,zhu2015co} which use all the frames, it loses some information. In this paper, our temporal attention module determines the degree of importance for each frame. Instead of skipping frames, it allocates different attention weights to different frames to automatically exploit their respective discriminative power and focus more on the important frames.
% for action recognition

\subsection{2.3~~Attention-Based Models}
%A human viewer would focus on some fixation points at the first glance of the scene,
When observing the real-world, a human usually focuses on some fixation points at the first glance of the scene, i.e., paying different attentions to different regions \cite{goferman2012context}. Many applications leverage predicted saliency maps for performance enhancement \cite{yu2010object,jiang2014saliency,bazzani2016recurrent}, which explicitly learn the saliency maps guided by human labeled groundtruths.
%,shapovalova2013action
% \cite {shapovalova2013action,bazzani2016recurrent}.
%When observing the real-world, a human usually focuses on some fixation points at the first glance of the scene, i.e., paying different attentions to different regions \cite{goferman2012context}. Many applications leverage predicted saliency map for performance enhancement, such as visual attention for robots \cite{yu2010object}, crowd analysis for video surveillance \cite{jiang2014saliency}, action recognition \cite {bazzani2016recurrent} and localization \cite{shapovalova2013action}. Those approaches explicitly learn the saliency maps guided by human labeled groundtruths.

The human labeled groundtruths for the explicit attention, however, are generally unavailable and might not be consistent with real attention related to the specific tasks.  Recently, the exploitation of an attention model which implicitly learns attention has attracted increasing interest in various fields, such as machine translation \cite{bahdanau2014neural}, image caption generation \cite{xu2015show}, and image recognition \cite{ba2014multiple}. Selective focus on different spatial regions is proposed for action recognition on RGB videos \cite{sharma2015action}. Ramanathan et al. propose an attention model which learns to detect events in RGB videos while attending to the people responsible for the event \cite{Ramanathan2015action}. The fusion of neighboring frames within a sliding window with learned attention weights is proposed to enhance the performance of dense labeling of actions in RGB videos \cite{yeung2015every}. However, all the attention models mentioned above for action recognition are based on RGB videos. There is a lack of investigation of skeleton sequences, which exhibit different characteristics from RGB videos.
%Selective focus on different spatial regions of the video frames is proposed for action recognition on RGB videos \cite{sharma2015action}.
%automatically

\section{3~~Overview of RNN and LSTM}

\begin{figure}[t]
	\centering
	\begin{subfigure}[t]{0.25\linewidth}
		\centering\includegraphics[width=\textwidth]{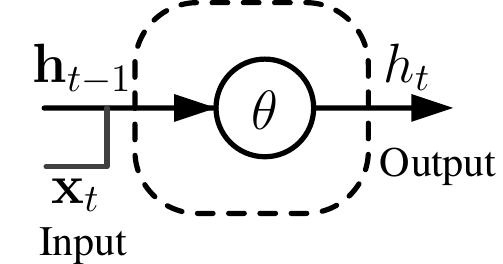}
		\caption{}
		\label{subfig:RNN}
	\end{subfigure}	
	\begin{subfigure}[t]{0.65\linewidth}
		\centering\includegraphics[width=\textwidth]{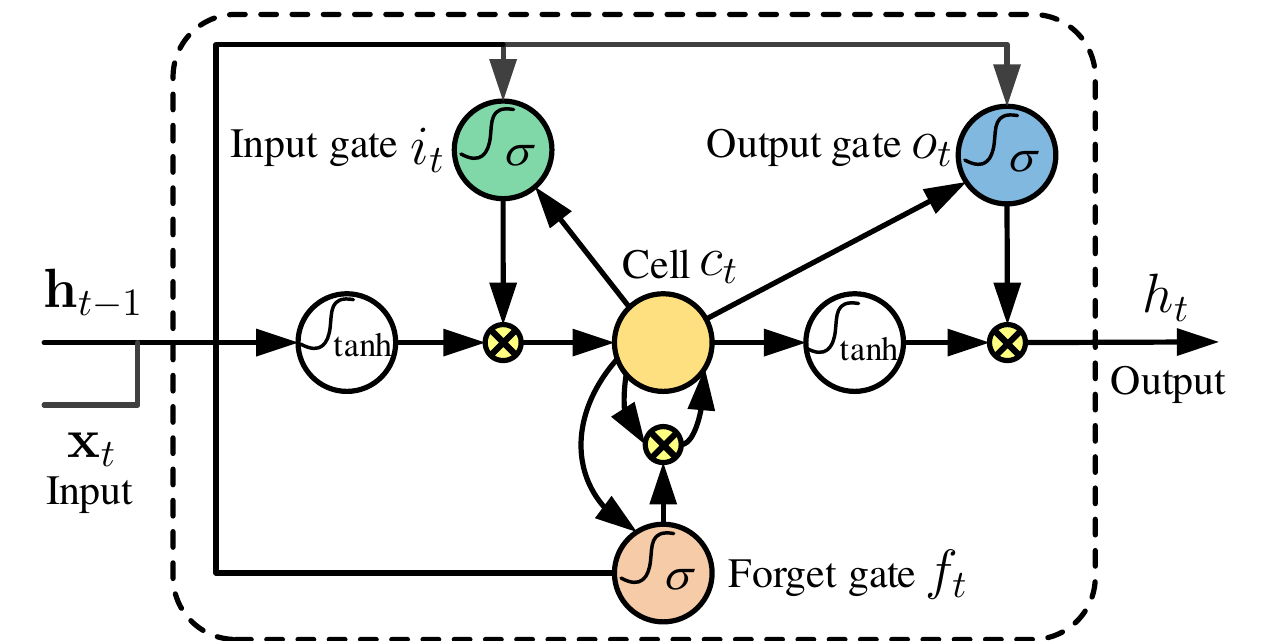}
		\caption{}			
		\label{subfig:LSTM_unit}
	\end{subfigure}
	\vspace{-3mm}
	\caption[]{Structures of the neurons. (a) RNN,
		(b) LSTM.}\label{fig:RNNLSTM}
	\vspace{1.5mm}
\end{figure}
In this section, we briefly review the Recurrent Neural Network (RNN), and Long Short-Term Memory (LSTM) to make the paper self-contained.

RNN is a popular model for sequential data modeling and feature extraction \cite{Graves2012}. Fig. \ref{fig:RNNLSTM}(a) shows an RNN neuron. The output response ${h}_t$ at time step $t$ is determined by the input $\mathbf{x}_t$ and the hidden outputs from RNN themselves at the last time step $\mathbf{h}_{t-1}$
%For an input sequence ${X} = \left( \mathbf{x}_1, ..., \mathbf{x}_T\right)$ with $T$ time steps, the output response ${h}_t$ at time step $t$ is determined by the input $\mathbf{x}_t$ and the hidden output from RNN at the last time step $\mathbf{h}_{t-1}$, which can be formulated as:
\begin{equation}
	\label{equ:RNN_ht}
	{h}_t = \theta \left( {\mathbf{w}}^\mathrm{T}_{xh} \mathbf{x}_t + {\mathbf{w}}^\mathrm{T}_{hh} \mathbf{h}_{t-1} + {b}_h \right),
\end{equation}
\begin{comment}
\begin{equation}
\label{equ:RNN_ht}
{h}_t = \theta \left( \mathsfsl{W}_{xh} \mathbf{x}_t + \mathsfsl{W}_{hh} \mathbf{h}_{t-1} + {b}_h \right),
\end{equation}
\end{comment}
where $\theta$ represents a non-linear activation function, $\mathbf{w}_{xh}$ and $\mathbf{w}_{hh}$ denote the learnable connection vectors, and ${b}_h$ is the bias value. The recurrent structure and the internal memory of RNN facilitate its modeling of the long-term temporal dynamics of the sequential data.
%$\mathsfsl{W}_{xh}$ denotes the connection weights from  the input layer to the hidden layer, $\mathsfsl{W}_{hh}$ denotes the recurrent weights from the hidden layer to itself, and ${b}_h$ is the bias value.
%at the two adjacent time steps

LSTM is an advanced RNN architecture which mitigates the vanishing gradient effect of RNN \cite{LSTM1997,vanish2001,Graves2012}. As illustrated in Fig. \ref{fig:RNNLSTM}(b), an LSTM neuron contains a memory cell ${c}_t$ which has a self-connected recurrent edge of weight 1. At each time step $t$, the neuron can choose to write, reset, and read the memory cell governed by the input gate ${i}_t$, forget gate ${f}_t$ and output gate ${o}_t$.

\section{4~~Deep LSTM with Spatio-Temporal Attention Model}
\label{sec:algorithm}
\begin{figure}[t] %{0.5\linewidth}
	\begin{center}
		\includegraphics[width=0.92\linewidth]{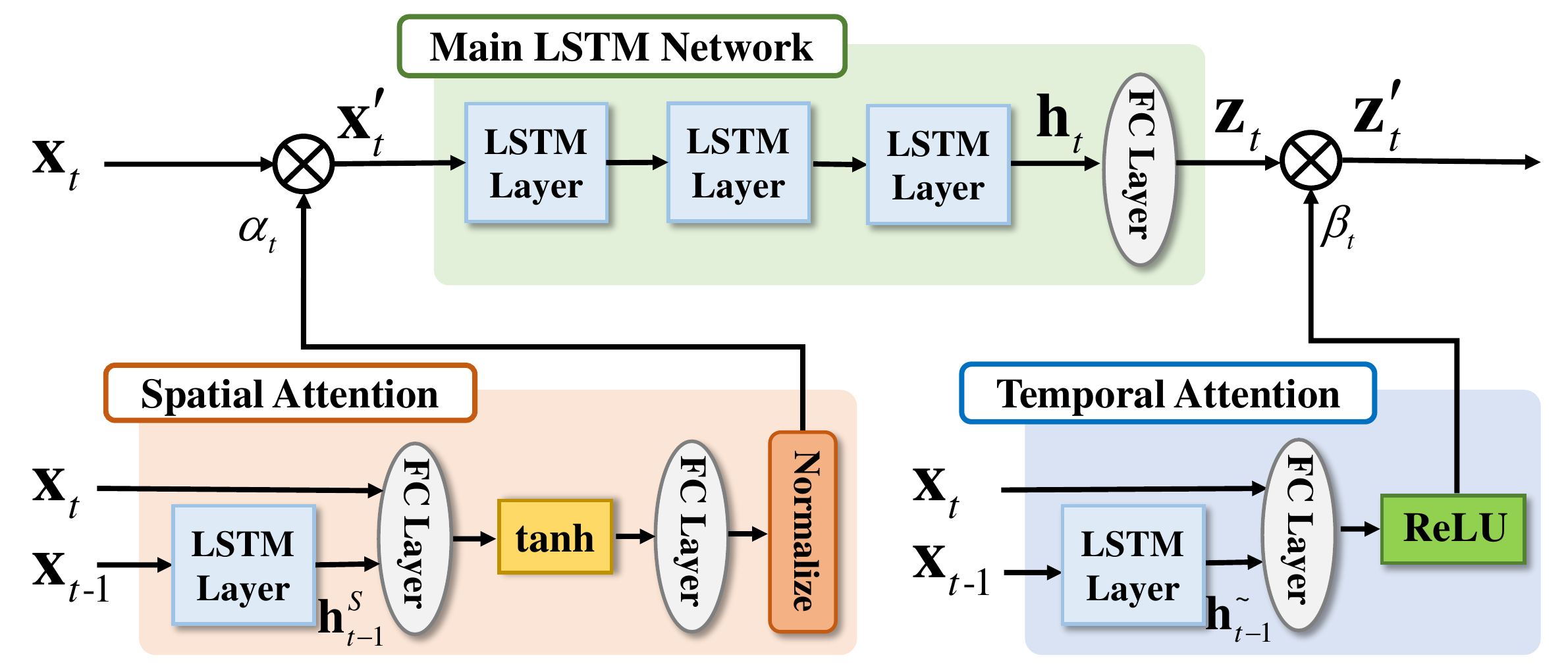}
	\end{center}
\vspace{-4mm}
	\caption{Overall architecture of our proposed network, which consists of the main LSTM network, the spatial attention subnetwork, and the temporal attention subnetwork.}
	\label{fig:Subnets}
\vspace{1.8mm}
\end{figure}
We propose an end-to-end multi-layered LSTM network with spatial and temporal attention mechanisms for action recognition. The network is designed to automatically select dominant joints within each frame through the spatial attention module, and assign different degrees of importance to different frames through the temporal attention module. Fig. \ref{fig:Subnets} shows its overall architecture, which consists of a main LSTM network, a spatial attention subnetwork, and a temporal attention subnetwork. Because of the inter-play among the three subnetworks, it is challenging to train the network.
%The input to the network is the 3D positions of joints of the skeletons.
%, considering spatial and temporal information simultaneously

In the following, we discuss the proposed spatial attention module and temporal attention module respectively, which are both built based on the LSTM networks. We then introduce a regularized learning objective of our model and a joint training strategy, which help overcome the difficulty of model learning for the highly coupled network.

%In the following of this section, we present our architecture of spatial-temporal attention model built based on the RNN model with LSTM as building blocks, with spatial attention modular and temporal attention modular discussed respectively. Finally, we will introduce a regularized learning objective of the our model with a two model alternate training strategy.

%In the following of this section, we first give a brief introduction of the RNN and LSTM network. Then we present our architecture of spatial-temporal attention model built based on the RNN model with LSTM as building blocks. After that, we respectively elaborate our spatial attention modular and temporal modular. Finally, we will introduce a regularized learning objective of the our model with a two model alternate training strategy.
%In this section, we first give a brief introduction of RNN and LSTM network. Then we introduce our attention-based network for action recognition with skeleton data.

\subsection{4.1 Spatial Attention with Joint-Selection Gates}

%Actions can be described with the coordinate information of skeleton data.
The action of persons can be described by the evolution of a series of human poses represented by the 3D coordinates of joints. In general, different actions involve different subsets of joints as discussed in Section 2.1. %For example, the joints ``arm", ``shoulder", and ``head" form the discriminative subset of joints for the action ``making telephone call". In contrast, the joints ``leg" and ``arm" are the more relevant joints for the action ``walking".
%%The performance of action recognition can be improved by mining discriminative actionlets and extracting combinatorial features \cite{CVPR12Actionlet}. The actionlets in \cite{CVPR12Actionlet} are mined from training set and consisted of fixed group of subsets of joints, but such attention is not sequence dependent. Moreover,  an action sequence may contain several substages and the active joints vary as the action proceeds. For example, a video sequence of the action ``shaking hands" may consist of several subactions such as ``approaching", ``shaking hands", and ``departing". Exploration of frame and content dependent attentions for different frames may better reflect the subactions and exclude noise.

We propose a spatial attention model to automatically explore and exploit the different degrees of importance of joints. With a soft attention mechanism, each joint within a frame is assigned a spatial attention weight based on the joint-selection gates. This enables our model to adaptively focus more on those discriminative joints.
% at each frame corresponding to the sequence

\begin{figure}[t]
	%\vspace{-6mm}
	\begin{center}
		\includegraphics[width=1\linewidth]{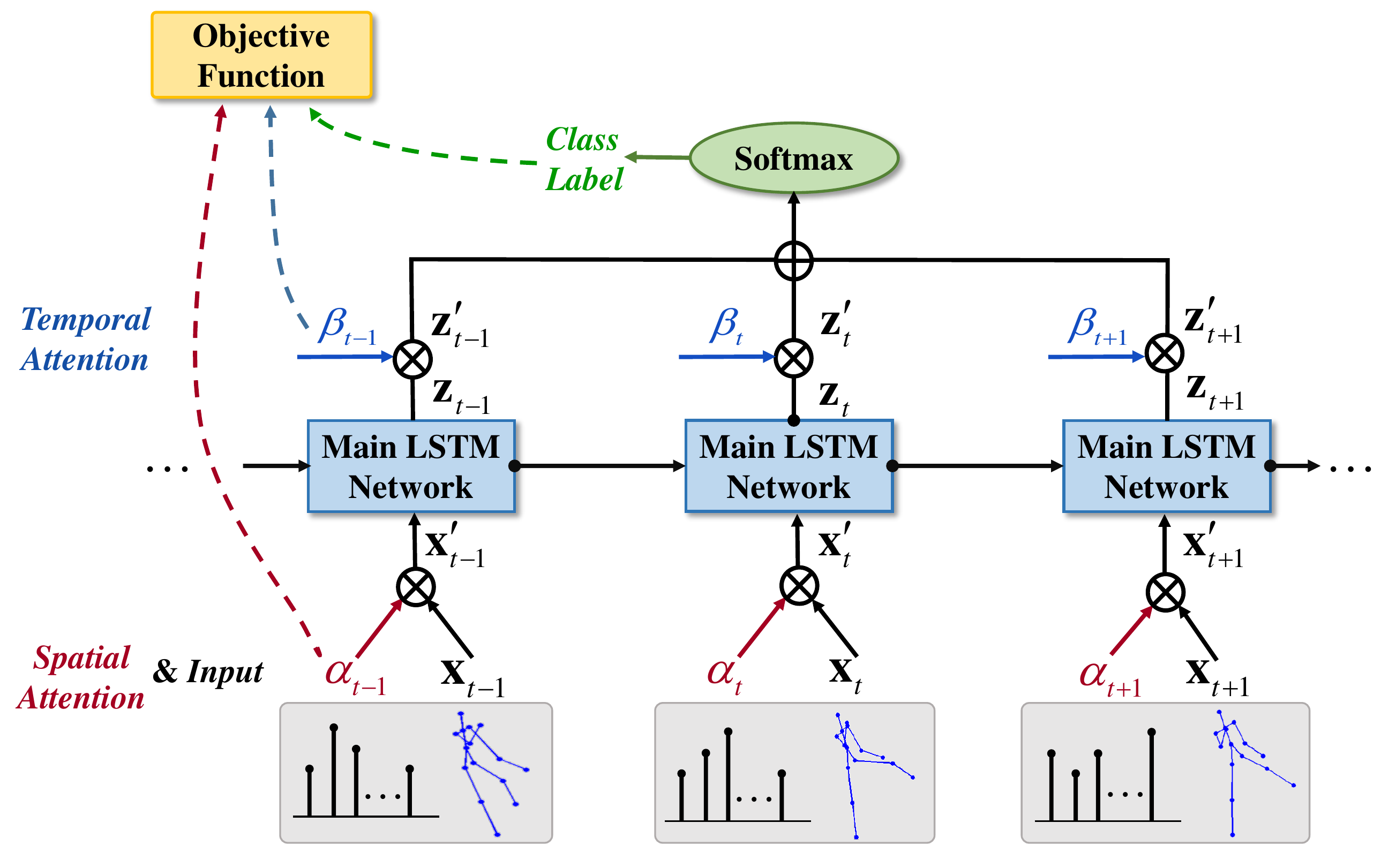} %framework.pdf}
	\end{center}
	\vspace{-4mm}
	\caption{Illustration of how spatial attention output $\mathbf{\alpha}$ and temporal attention output $\beta$ influence the LSTM network.}
	%The network of our proposed end-to-end spatio-temporal attention model (need to modify later .......
	\label{fig:network}
	\vspace{1mm}
\end{figure}

At each time step $t$, given the full set of $K$ joints $\mathbf{x}_t = \left( \mathbf{x}_{t,1}, ..., \mathbf{x}_{t,K} \right)^\mathrm{T}$, with $\mathbf{x}_{t,k}\in \mathbb{R}^{3}$, the scores $\mathbf{s}_t=(s_{t,1}, \cdots, s_{t,K})^\mathrm{T}$ for indicating the importance of the $K$ joints are jointly obtained as
%in $\Re^K$
\begin{equation}
\label{equ:spa_att_2}
\mathbf{s}_t = \mathsfsl{U}_{s} \tanh(\mathsfsl{W}_{xs}\mathbf{x}_t + \mathsfsl{W}_{hs}\mathbf{h}_{t-1}^{s} + \mathbf{b}_{s}) + \mathbf{b}_{us},
% \mu ( \mathbf{X}_t, \mathbf{h}_{t-1}, \Theta_s),
\end{equation}
where $\mathsfsl{U}_s$, $\mathsfsl{W}_{xs}$, $\mathsfsl{W}_{hs}$ are the learnable parameter matrixes, $\mathbf{b}_{s}$, $\mathbf{b}_{us}$ are the bias vectors. $\mathbf{h}_{t-1}^{s}$ denotes the hidden variable from an LSTM layer as illustrated in Fig. \ref{fig:Subnets}. For the $k^{th}$ joint, the activation as the joint-selection gate is computed as
%The model considers the hidden variables $\mathbf{h}_{t-1}^{s}$ from a LSTM layer, and all the joints $\mathbf{x}_t$ at the current time step $t$.
\begin{equation}
\label{equ:spa_att_1}
\alpha_{t,k} = \frac{\exp(s_{t,k})}{\sum_{i=1} ^ K \exp(s_{t,i})},
\end{equation}
which is a normalization of the scores. The set of gates controls the amount of information of each joint to flow to the main LSTM network. Among the joints, the larger the activation, the more important this joint is for determining the type of action. We also refer to the activation values as attention weights. Instead of assigning equal degrees of importance to all the joints $\mathbf{x}_t$, as illustrated in Fig. \ref{fig:network}, the input to the main LSTM network is modulated to $\mathbf{x'}_t = \left( \mathbf{x'}_{t,1}, ..., \mathbf{x'}_{t,K} \right)^\mathrm{T}$, with $\mathbf{x'}_{t,k} = \alpha_{t,k} \cdot \mathbf{x}_{t,k}$.
\begin{comment}
\begin{equation}
\label{equ:spa_att_con}
\mathbf{x'}_{t,k} = \alpha_{t,k} \cdot \mathbf{x}_{t,k}.
\end{equation}
\end{comment}

Note that the proposed spatial attention model determines the importance of joints based on all the joints of the current time step and the hidden variables from an LSTM layer. On one hand, the hidden variables $\mathbf{h}_{t-1}$ contain information of past frames, benefiting from the merit of LSTM which is capable of exploring temporal long range dynamics. In this paper, the spatial attention subnetwork composes of an LSTM layer, two fully connected layers and a normalization unit as illustrated in Fig. \ref{fig:Subnets}. On the other hand, leveraging all joints within the current frame provides necessary ingredient for determining their importance.
%LSTM subnetwork of the spatial attention module  is a one layer network with multiple (i.e., $N_s$) LSTM neurons as illustrated by the spatial attention module in Fig. \ref{fig:Subnets}.

Bridged by the joint-selection gate, the main LSTM network and the spatial attention subnetwork can be jointly trained to implicitly learn the spatial attention model.

\subsection{4.2 Temporal Attention with Frame-Selection Gate}

%In LSTM network, the output vectors of hidden layers form the frame-wise features which also model spatial and temporal information.
For a sequence, the amount of valuable information provided by different frames is in general not equal. Only some of the frames (key frames) contain the most discriminative information while the other frames provide context information. For example, for the action ``shaking hands", the sub-stage ``approaching" should have lower importance than the sub-stage of ``hands together". Based on such insight, we design a temporal attention module to automatically pay different levels of attention $\beta$ to different frames.

%For the sequence level classification, based on the hidden output $\mathbf{z}_t$ of the main LSTM network and the temporal attention value $\beta_t$ at each time step $t$ as illustrated in Fig. \ref{fig:network}, the probability that a sequence $X$ belongs to class $k$ is
For the sequence level classification, based on the output $\mathbf{z}_t$ of the main LSTM network and the temporal attention value $\beta_t$ at each time step $t$, the scores for $C$ classes are the weighted summation of the scores at all time steps
\vspace{-1mm}
\begin{equation}
	\label{equ:ot}
\vspace{-1mm}
	\mathbf{o} = \sum_{t=1}^{T}{\beta_t} \cdot \mathbf{z}_t,
\vspace{-0mm}
\end{equation}
where $\mathbf{o} = (o_1, o_2, \cdots, o_C)^\mathrm{T}$, $T$ denotes the length of the sequence. Fig. \ref{fig:network} illustrates how the temporal attention output $\beta$ is incorporated to the main LSTM network. The predicted probability being the $i^{th}$ class given a sequence $X$ is
\begin{equation}
\vspace{-2mm}
\label{equ:pred}
p\left( C_i | \mathsfsl{X} \right) = \frac{e^{o_i}}{\sum_{j=1}^C e^{o_j}},~~k = 1,..., C.
\end{equation}
%where $T$ denotes the length of the sequence. $\mathbf{o} = (o_1, o_2, \cdots, o_C)^\mathrm{T}$ represents the fused final scores for $C$ classes. It is the weighted summation of the scores $\mathbf{z}_t, t = 1, \cdots, T,$ with weights from the learned temporal attentions across the $T$ time slots. Fig. \ref{fig:network} gives an illustration on how the temporal attention output $\beta$ is introduced to the main LSTM network.
% as illustrated in Fig. \ref{fig:network}

As illustrated in Fig. \ref{fig:Subnets}, the attention module is composed of an LSTM layer, a fully connected layer, and a ReLU non-linear unit, being connected in series. It plays the role of soft frame selection. The activation as the frame-selection gate can be computed as
%with soft frame-selection gates
\begin{equation}
\label{equ:tem_att}
\beta_t = \text{ReLU}(\mathbf{w}_{x\thicksim}\mathbf{x}_t + \mathbf{w}_{h\thicksim}\mathbf{h}_{t-1}^{\thicksim} + b_{\thicksim}),
%\beta_t = \text{ReLU}(\mathbf{w}_{x\}\mathbf{x}_t + \mathbf{w}_{h\thicksim}\mathbf{h}_{t-1} + b_{\thicksim}),
\end{equation}
which depends on the current input $\mathbf{x}_t$, and the hidden variables $\mathbf{h}_{t-1}^{\thicksim}$ of time step $t-1$ from an LSTM layer. We use the non-linear function of ReLU due to its good convergence performance. The gate controls the amount of information of each frame to be used for making the final classification decision. The works \cite{CVPR15HRNN,zhu2015co} are our special cases where the attention weights on each frame are equal.

Bridged by the frame-selection gate, the main LSTM network and the temporal attention subnetwork can be jointly trained to implicitly learn the temporal attention model.

\subsection{4.3 Joint Spatial and Temporal Attention}

The purpose of the attention models is to enable the network to pay different levels of attention to different joints and assign different degrees of importance to different frames as an action proceeds. We integrate spatial and temporal attention in the same network as illustrated in Fig. \ref{fig:Subnets}. How the spatial attention model acts on the input and how the temporal attention model acts on the output of the main LSTM network are illustrated in Fig. \ref{fig:network}.

\subsubsection{Regularized Objective Function}

We formulate the final objective function of the spatio-temporal attention network with a regularized cross-entropy loss for a sequence as,
\begin{equation}
\label{equ:lossfuntion}
\vspace{-2mm}
\begin{split}
L = &-\sum_{i=1}^C y_i\log \hat{y}_i + \lambda_1 \sum_{k=1}^K \left( 1 - \frac{\sum_{t=1}^T \alpha_{t,k}}{T} \right)^2 \\
\vspace{-2mm}
&+  \frac{ \lambda_2}{T}\sum_{t=1}^T \|\beta_t\|_2
 + \lambda_3 \| \mathsfsl{W}_{u v}\|_1,
%\sum_i \sum_j {\theta}_{i,j}^2,
\end{split}
\vspace{-0mm}
\end{equation}
where $\mathbf{y} = (y_1, \cdots, y_C)^\mathrm{T}$  denotes the groundtruth label. If it belongs to the $i^{th}$ class, then $y_i\!=\!1$ and $y_{j}\!=\!0$ for $j\!\neq\!i$. $\hat{y_i}$ indicates the probability that the sequence is predicted as the $i^{th}$ class, where $\hat{y}_{i}\!=\!p (C_i|X)$. The scalars $\lambda_1$, $\lambda_2$, and $\lambda_3$ balance the contribution of the three regularization terms. We discuss the regularization designs in the following.  %We add the $l_1$ norm regularization which makes a kind of compromise between finding small weights and minimizing the first two items to reduce overfitting.

The first regularization item is designed to encourage the spatial attention model to dynamically focus on more spatial joints  in a sequence. We found the spatial attention model is prone to consistently ignoring many joints along time even though these joints are also valuable for determining the type of action, i.e., trapped to a local optimum. We introduce this regularization item to avoid such ill-posed solutions. For clarity, we re-describe it as $\sum_{t=1}^T \alpha_{t,k} \!\approx\!T$, with $k = 1,\cdots, K$. This encourages paying equal attentions to different joints.
%There is an implicit constraint for all joints at each time slot $t$, $\sum_{k=1}^K \alpha_{t,k} = 1$ as indicated by (\ref{equ:spa_att_1}).
\begin{comment}
\begin{equation}
\label{equ:spa_constraint}
\sum_{t=1}^T \alpha_{t,k} \approx T.
\end{equation}
\end{comment}

The second regularization item is to regularize the learned temporal attention values under control with $l_2$ norm rather than to increase them unboundedly. This alleviates  gradient vanishing in the back propagation, where the back-propagated gradient is proportional to $1/\beta_{t}$.

The third regularization item with $l_1$ norm is to reduce overfitting of the networks. $\mathsfsl{W}_{u v}$ denotes the connection matrix (merged to one matrix here) in the networks.
%, as well as the attention networks.

%\subsection{4.4~~ Combination of Spatial and Temporal Attention and Joint Training}

\subsubsection{Joint Training of the Networks} Due to the mutual influence of the three networks, the optimization is rather difficult. We propose a joint training strategy to efficiently train the spatial and temporal attention modules, as well as the main LSTM network. The separate pre-training of the attention modules ensures the convergence of the networks. The training procedure is described in Algorithm \ref{alg:Framwork}.
%Due to the mutual influence of the spatial attention model and temporal attention model to each other and to the entire network,
%We propose to alternatively fix the spatial attention subnetwork and the temporal attention subnetwork for the training. This is inspired by the policy network from AlphaGo \cite{silver2016mastering}, where self plays with an old, fixed opponent and update the old opponent every 500 iteration rather than update the self and the opponent simultaneously. This is beneficial for achieving stable training. We describe the training procedure in Algorithm \ref{alg:Framwork}.
\vspace{-3mm}
\begin{algorithm}[htb]
	%\caption{Combined training of spatial and temporal attention based network.}

	\caption{Joint Training of the LSTM Network with Spatio-Temporal Attention Model.}
	\label{alg:Framwork}
	\begin{algorithmic}[1]
	    \REQUIRE model training parameters $N_1$, $N_2$ (e.g., $N_1=1000$, $N_2=500$).
	    % used in the experiments
	
	    \STATE Initialize the network parameters using Gaussian. %XIVIAR or
	%Initialization: initialize the model parameters using XIVIAR or Gasussian.
	
	    \textbf{// Pre-train Temporal Attention Model.}
	     	
		\STATE With spatial attention weights being fixed as ones, jointly train the main LSTM network with only one LSTM layer and the temporal attention subnetwork to obtain the temporal attention model.
		\STATE Fix this learned temporal attention subnetwork. Train the main LSTM network after increasing its number of LSTM layers to three by $N_1$ iterations.
		\STATE Fine-tune this temporal attention subnetwork and the main LSTM network by $N_2$ iteration.
		
		\textbf{// Pre-train Spatial Attention Model.}
		
		\STATE With temporal attention weights being fixed as ones, jointly train the main LSTM network with only one LSTM layer and the spatial attention subnetwork to obtain the spatial attention model.
		\STATE Fix this learned spatial attention subnetwork. Train the main LSTM network after increasing its number of LSTM layers to three by $N_1$ iterations.
		\STATE Fine-tune this spatial attention subnetwork and the main LSTM network for $N_2$ iterations.
		
		\textbf{// Train the Main LSTM Network.}
		
		\STATE Fix both the temporal and spatial attention subnetworks learned in Step-4 and Step-7. Fine-tune the main LSTM network by $N_1$ iterations.
		
		\textbf{// Jointly Train the Whole Network.}
		\STATE Jointly fine-tune the whole network (main LSTM network, the spatial attention subnetwork, and the temporal attention subnetwork) by $N_2$ iterations.
		\ENSURE  the final converged whole model.
	\end{algorithmic}
\end{algorithm}
\vspace{-3mm}

\section{5.~~Experimental Results}
%In this section, we present the experimental analyses and evaluations of the proposed models and other approaches. We will discuss the datasets and settings, visualization of the learned attentions, analysis of the proposed models, and the overall results respectively below.
%For CMU dataset, we use the data labelled by \cite{zhu2015co}.
% on two datasets: SBU kinect interaction dataset \cite{yun2012two}, and the largest RGB+D dataset of NTU \cite{Shahroudy_2016_CVPR}
\subsection{5.1~~Datasets and Settings}
We perform our experiments on the following two datasets: the SBU Kinect interaction dataset \cite{yun2012two}, and the largest RGB+D dataset of NTU (Shahroudy et al. 2016).

\textbf{SBU Kinect Interaction Dataset (SBU).} The SBU dataset is an interaction dataset with two subjects. It contains 230 sequences of 8 classes (6614 frames) with subject independent 5-fold cross validation. Each person has 15 joints and the dimension of the input vector is $15 \times 3 \times 2 = 90$. Note that we smooth each joint's position of the skeleton in the temporal domain to reduce the influence of noise \cite{CVPR15HRNN,zhu2015co}.
%\textbf{CMU Dataset.} There are 2235 sequences (987,431 frames) with 45 classes in the entire dataset while the subset contains 664 sequences (125,667 frames) with 8 classes. The video lengths are diverse and action may repeat within one video. We adopt three-fold cross validation to report the performance as in \cite{zhu2015co}.
%Besides, it is pretty challenging to handle intra-class variations.

\textbf{NTU RGB+D Dataset (NTU).} The NTU dataset is currently the largest action recognition dataset with high quality skeleton \cite{Shahroudy_2016_CVPR}. It contains 56880 sequences (with 4 million frames) of 60 classes, including Cross-Subject (CS) and Cross-View (CV) settings. Each person has 25 joints. We apply the similar normalization preprocessing step to have position and view invariance \cite{Shahroudy_2016_CVPR}. To avoid destroying the continuity of a sequence, no temporal down-sampling is performed.
%As the number of actors in different frames is not consistent, we fix the number of people in a video to be the maximum person number of this dataset.
%, with the empty person filled by zeros
%, including 40 daily actions (drinking, eating, reading, etc.), 9 health-related actions (sneezing, staggering, falling down, etc.) and 11 mutual action (punching, kicking, hugging, etc.)
%Note that for all the datasets, we normalize the data to exclude noises and handle the view variation \cite{CVPR15HRNN,zhu2015co}.

\textbf{Implementation Details.} For the network and parameter settings, we use three LSTM layers for the main LSTM network, and one LSTM layer for each attention network. Each LSTM layer composes of 100 LSTM neurons. We set $\lambda_1$, $\lambda_2$, and $\lambda_3$ to $0.001$, $0.0001$, and $0.0005$ for the SBU dataset, and $0.01$, $0.001$ and $0.00005$ for the NTU dataset experimentally. Adam \cite{kingma2014adam} is adopted to automatically adjust the learning rate during optimization. The batch sizes for the SBU dataset and the NTU dataset are $8$ and $256$ respectively. Dropout is utilized to mitigate overfitting \cite{ICLR15DropoutLSTM}.
%The initial learning rate is 0.0002 and 0.002 for SBU Dataset and NTU Dataset, respectively.
%The number of neurons is set to $100$ for each layer.

%The raw data is normalized and smoothed by the Svaitzky-Golay filter in the temporal domain \cite{CVPR15HRNN,zhu2015co} to reduce noises.

\subsection{5.2~~Visualization of the Learned Attentions}
We analyze where the learned spatial and temporal attention attend to by visualizing the attention weights in the test.
\begin{figure}[http]
	\vspace{-3mm}
	\centering
	\begin{subfigure}[t]{0.36\textwidth}
		\centering\includegraphics[scale=0.28]{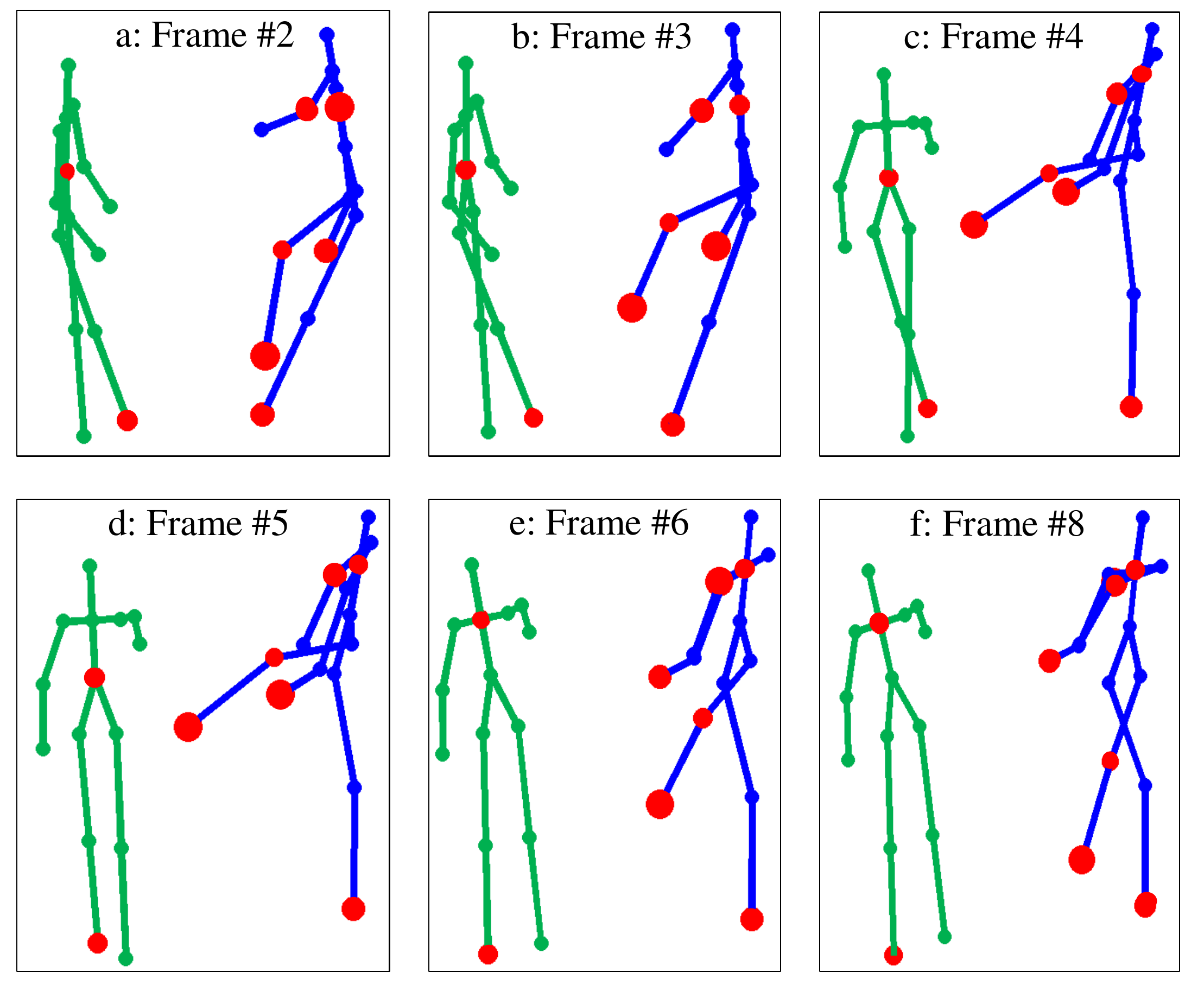}
		\vspace{-2mm}
		\caption{}
		\label{fig:vis_tem_skeleton}
	\end{subfigure}
	
	\begin{subfigure}[t]{0.22\textwidth}
		\centering\includegraphics[scale=0.27]{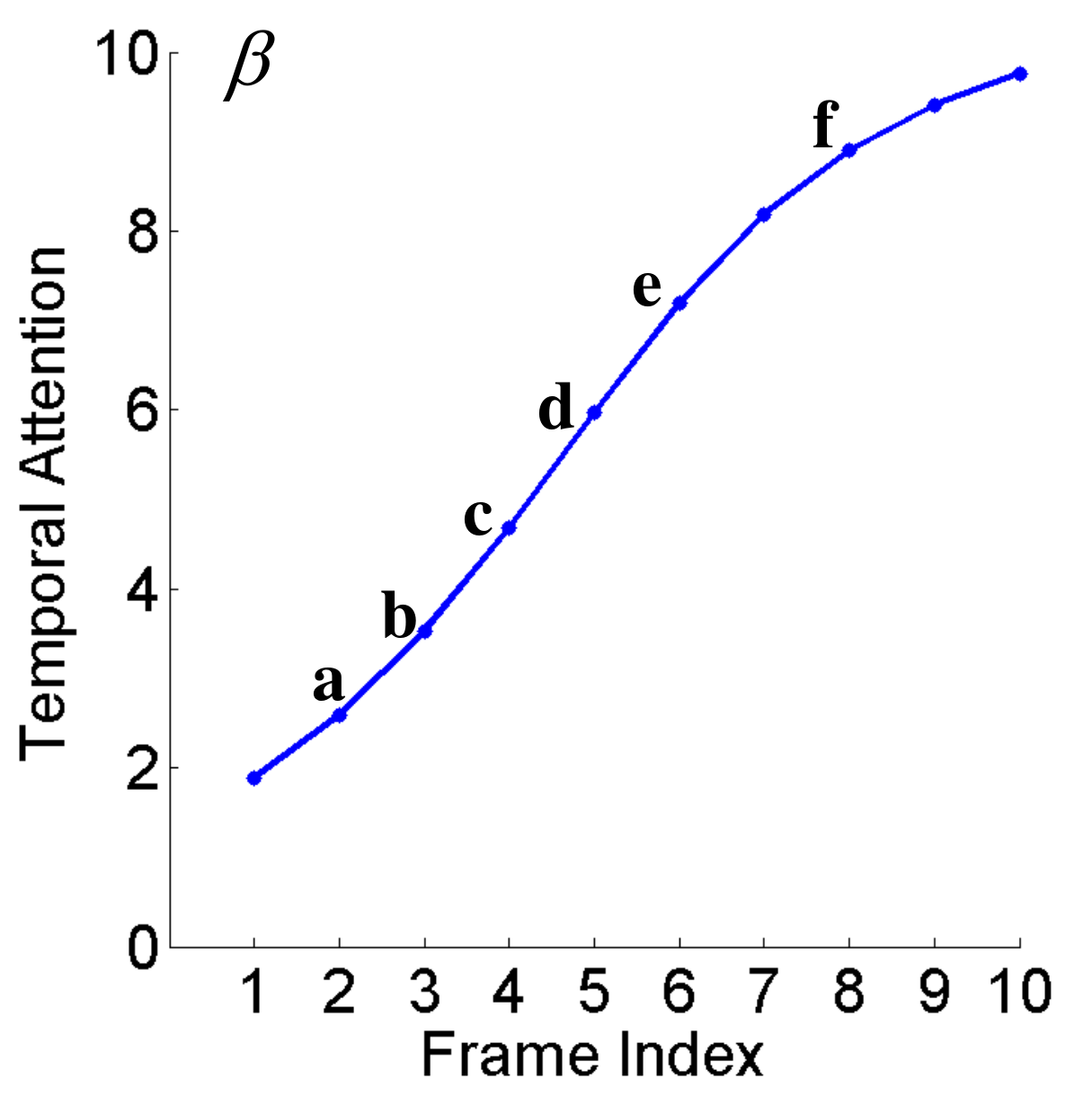}
		\vspace{-2.2mm}
		\caption{}
		\label{fig:skeleton2}
	\end{subfigure}
	\begin{subfigure}[t]{0.22\textwidth}
		\centering\includegraphics[scale=0.28]{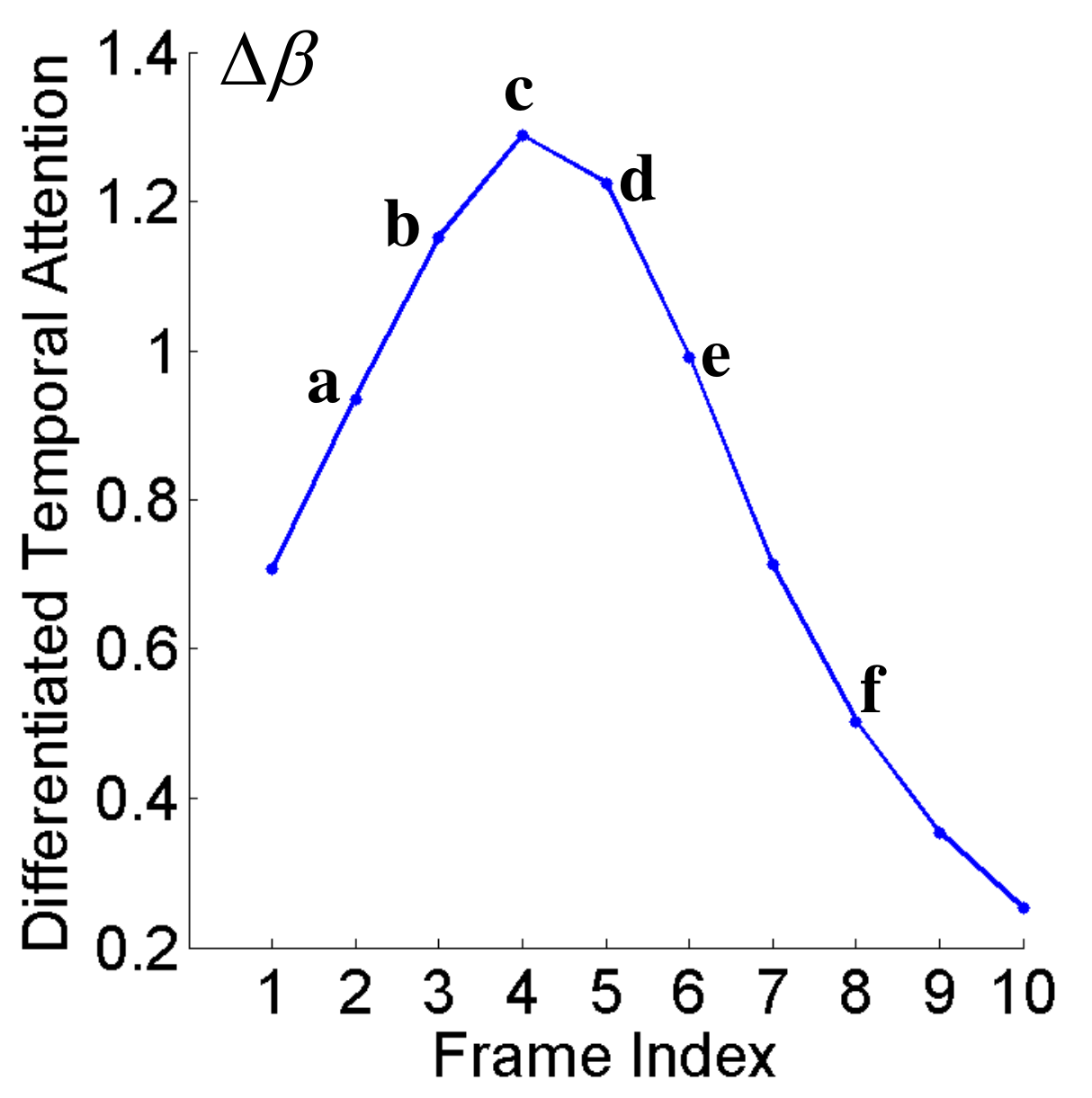}
		\vspace{-2.2mm}
		\caption{}
		\label{fig:skeleton2}
	\end{subfigure}
	\vspace{-4mm}
	\caption[]{Visualization of the spatial and temporal attention weights from our model for the action ``kicking". (a) Spatial attention weights. The larger of the red circle, the higher of the attention on that joint. We only mark on the 8 joints with the largest attentions. (b) Temporal attention weights $\beta$ on each frames. (c) Differentiated temporal attention weights (i.e., $\vartriangle\beta_t = \beta_t - \beta_{t-1}$). Best viewed in color.}\label{fig:discuss}
\end{figure}
\vspace{-2mm}

\begin{comment}
\begin{figure}[th]
	\vspace{-3mm}
	\begin{center}
		\includegraphics[scale = 0.17]{figs/vis_tem.png}
	\end{center}
	\vspace{-4mm}
	\caption{Visualization of spatial and temporal attention weights of ``Kicking" (top) and ``Pushing" (bottom). (a) to (f) show the spatial attention weights (marked by green circles, the larger of the circle, the higher of the attention intensity) on those frames. The right figure shows the temporal attention weights as the action proceeds. Best viewed in color.}
	\label{fig:discuss}
	\vspace{0mm}	
\end{figure}
\end{comment}

% indicate the relative amplitudes of the spatial attention for each joint
%as the action proceeds
%\textbf{Spatial Attention.} For action ``Kicking", Fig. \ref{fig:discuss} (a) to (f) show the variations of spatial attention weights from the learned model. The color and circle size of a joint indicate the amplitude of the spatial attention weight of that joint. The attention weight for the right foot of the right person is increasing as he stretches his leg. In the meanwhile, the weight on the torso of the left person is growing as he is falling backward. We show the spatial attention weights at the sequence level in Fig. \ref{fig:spa_att_vis}. From the perspective of the whole sequence, the right foot and the left arm of the right person, the torso and the right foot of the left actor are the most discriminative joints for the action ``Kicking". The learned important joints as the action proceeds are consistent with what human perceives.

\begin{figure}[t]
	\begin{center}
		\includegraphics[width=0.85\linewidth]{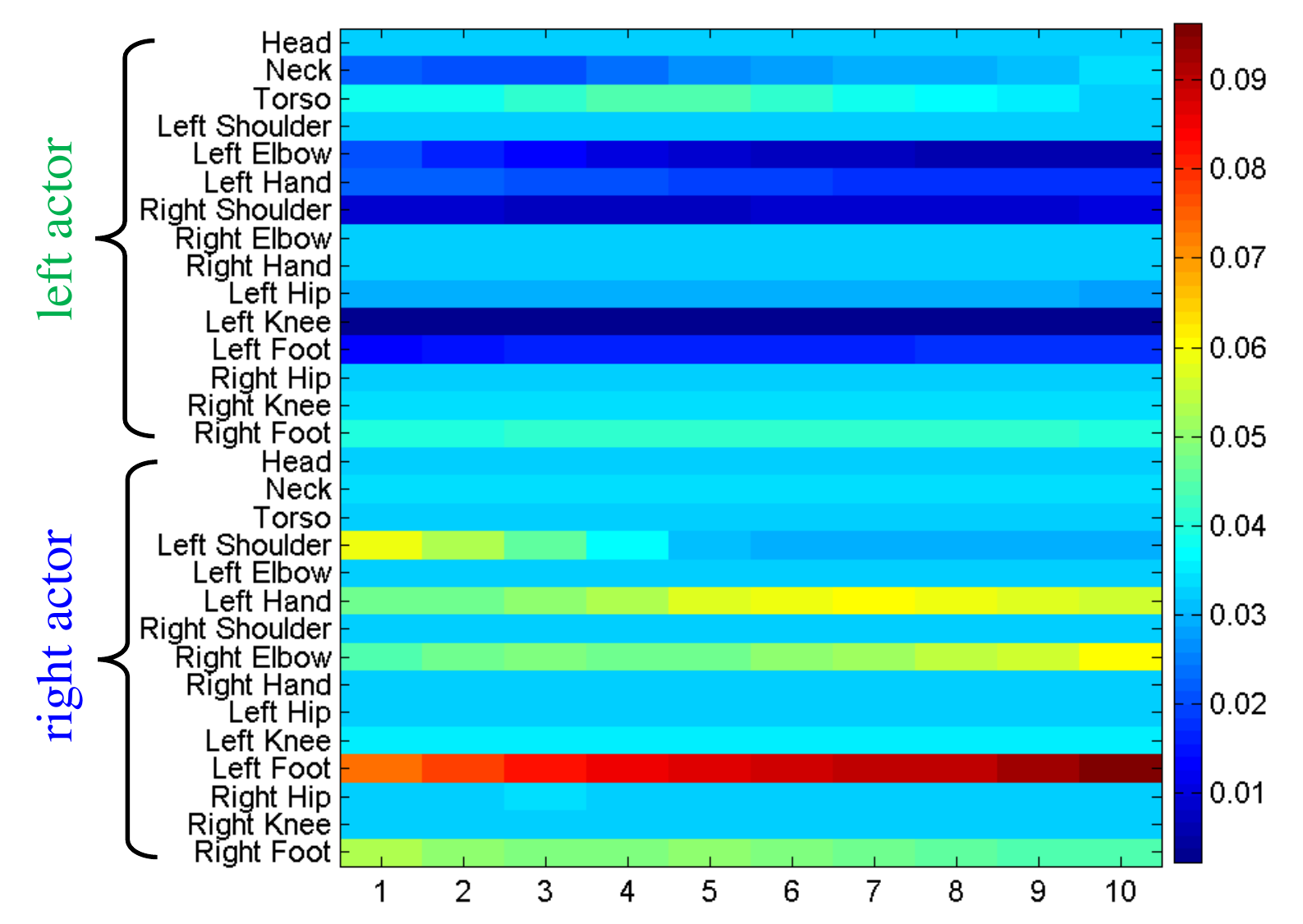}
\vspace{-4mm}
	\end{center}
	%\caption{The visualization of spatial attention weight on the two actors of the action ``Punching" for a sequence.  (a) The map for joint positions and indexes(a). (b) Horizontal axis denotes the joint indexes. Vertical axis denotes the frame id (time). The color values indicate the spatial attention weights.}
	\caption{Visualization of spatial attention on the two actors of the action ``kicking" for a sequence. Vertical axis denotes the joint indexes. Horizontal axis denotes the frame indexes (time). Color values indicate the spatial attention weights.}
	% The joints for the left actor are in the upper half while those for the right actor are in the bottom half.
	\label{fig:spa_att_vis}
\vspace{1mm}
\end{figure}

\textbf{Spatial Attention.} For a sequence of action ``kicking", Fig. \ref{fig:discuss}(a) shows the amplitude of the spatial attention weights on joints by the sizes of the red circles. We also present concrete attention values in Fig. \ref{fig:spa_att_vis}. The attention weights on the left foot, right elbow and left hand of the right person are large. Meanwhile, the weights on the torso and right foot of the left person are large. Being content-dependent, the attentions vary across frames. The learned important types of joints are consistent with what human perceives.
%From the perspective of the whole sequence, the left foot, left hand, and the right elbow of the right person, the torso and the right foot of the left person are the most discriminative joints for this sequence of action ``Kicking".

\textbf{Temporal Attention.} Fig. \ref{fig:discuss}(b) shows the temporal attention weights $\beta$. Fig. \ref{fig:discuss}(c) shows the differentiated attention weights (i.e., $\vartriangle\!\!\beta_t = \beta_t - \beta_{t-1}$) for ``Kicking". Since the LSTM network usually accumulates more information as time goes, the attention weight usually increases correspondingly. The increased amplitude of the attention weight, i.e., $\vartriangle\!\!\beta_t$, can indicate the importance of the frame $t$. We can see the differentiated attention weight goes up to a climax as the person on the right lifts his foot to the highest point, which human also considers as more discriminative.

\subsection{5.3~~Effectiveness of the Proposed Attention Models}
\begin{figure}[th]
	\vspace{-3mm}
	\centering
	\begin{subfigure}[t]{0.15\textwidth}
		\centering\includegraphics[width=2.7cm,height=3.5cm]{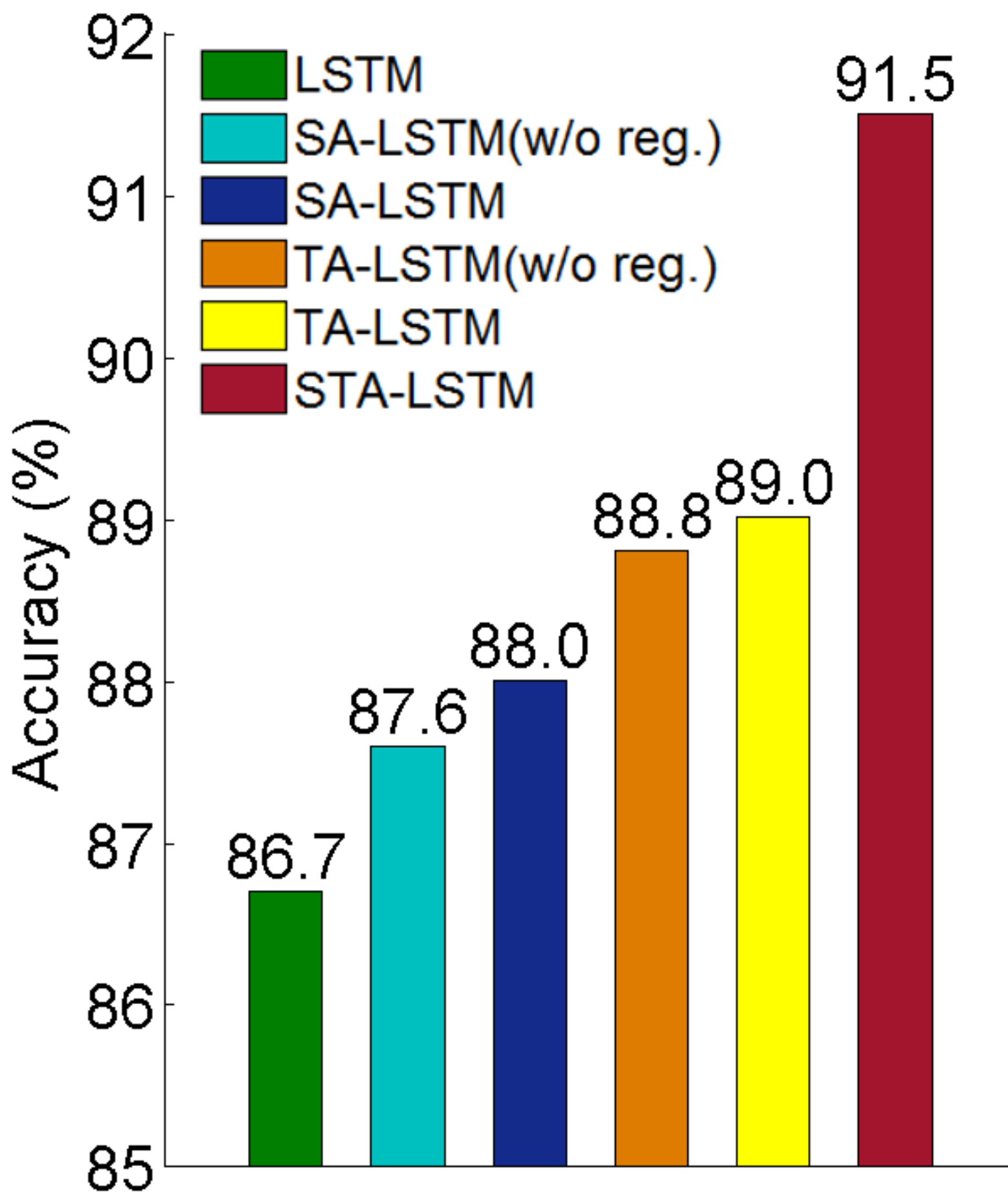}
		\vspace{-4mm}
		\caption{SBU}
		\label{fig:skeleton1}
		
	\end{subfigure}
	\begin{subfigure}[t]{0.15\textwidth}
		\centering\includegraphics[width=2.7cm,height=3.5cm]{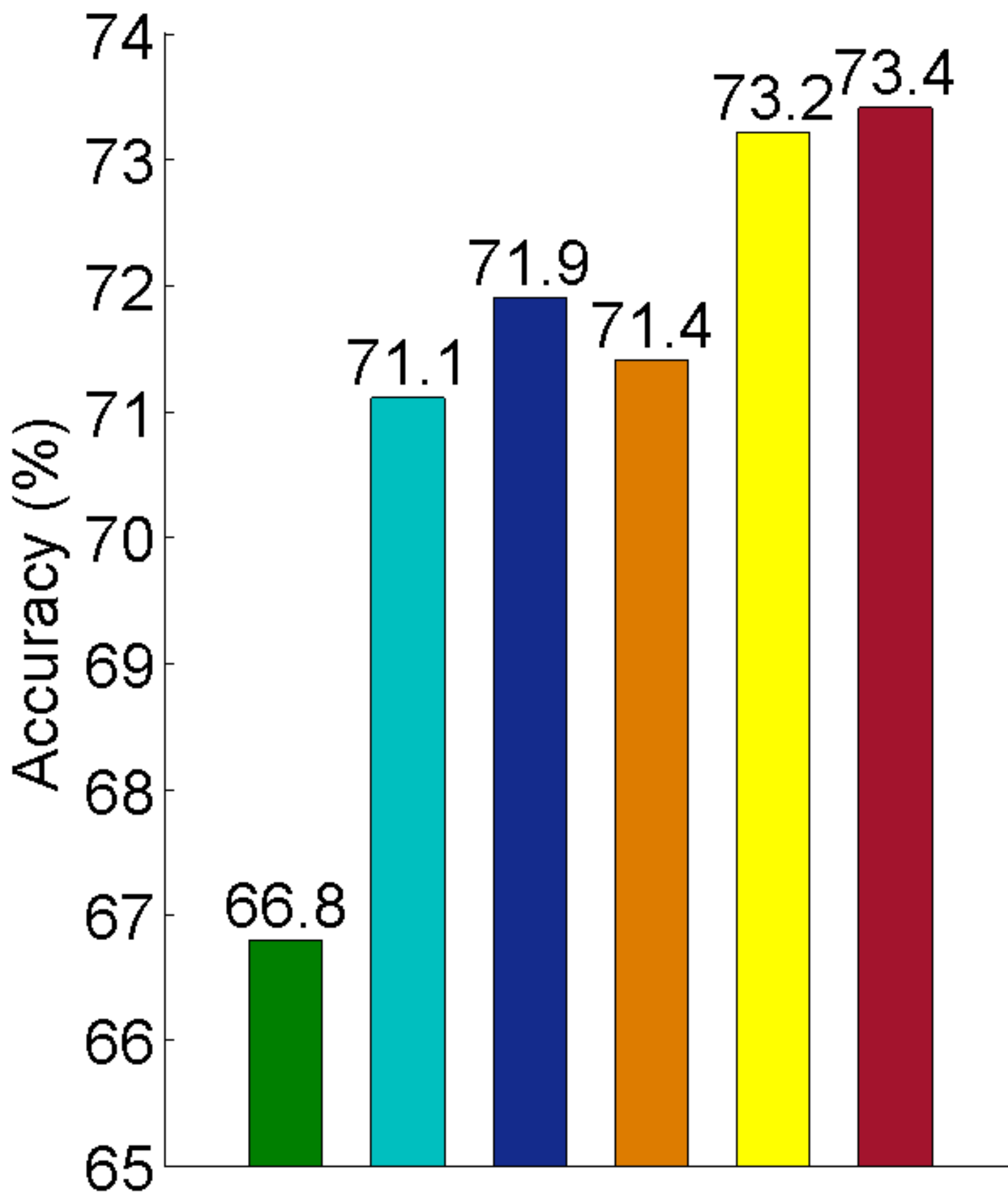}
		%\vspace{-2mm}
		\vspace{-4mm}
		\caption{NTU-CS}
		\label{fig:skeleton2}
		
	\end{subfigure}
	\begin{subfigure}[t]{0.15\textwidth}
		\centering\includegraphics[width=2.7cm,height=3.5cm]{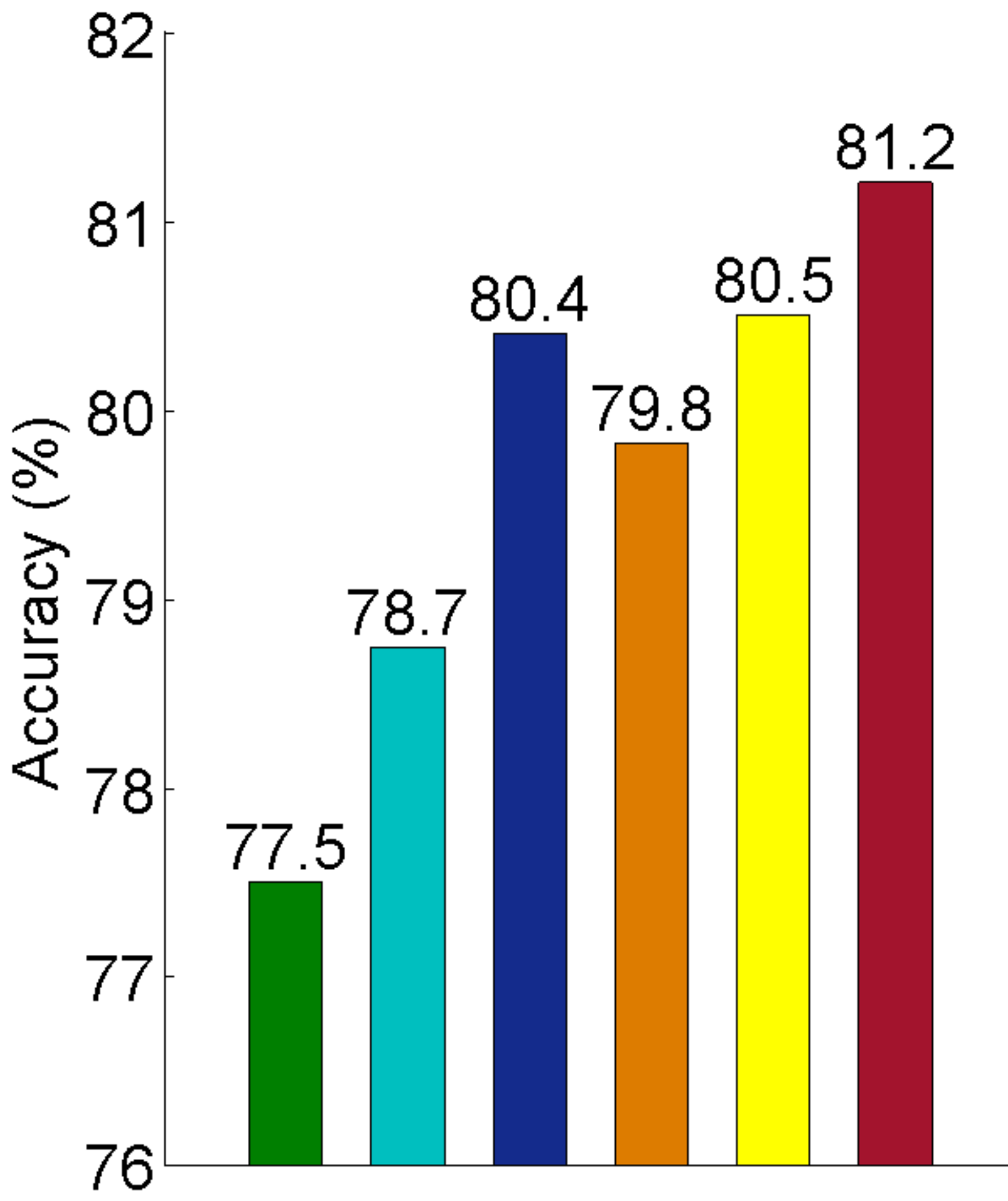}
		\vspace{-4mm}
		\caption{NTU-CV}
		\label{fig:skeleton2}
	\end{subfigure}
	\vspace{-3mm}
	\caption[]{Performance evaluation of our attention models, and the regularization items on two datasets in accuracy (\%).}\label{fig:in-comp}
\vspace{-3.5mm}	
\end{figure}
\vspace{1.8mm}
To validate the effectiveness of our designs, we conduct experiments with different configurations as follows.
\vspace{-1mm}
\begin{itemize}
\setlength{\itemsep}{0pt}
\setlength{\parskip}{0pt}
	\item \textbf{LSTM}: main LSTM network without attention designs.
	\item \textbf{SA-LSTM(w/o reg.)}: LSTM  + spatial attention without regularization (only includes $1^{st}$ and $4^{th}$ items in (\ref{equ:lossfuntion})).
	\item \textbf{SA-LSTM}: LSTM  + spatial attention network.
	\item \textbf{TA-LSTM(w/o reg.)}: LSTM + temporal attention without regularization(only includes $1^{st}$ and $4^{th}$ items in (\ref{equ:lossfuntion})).
	\item \textbf{TA-LSTM}: LSTM + temporal attention network.
	\item \textbf{STA-LSTM}: LSTM+spatio-temporal attention network.
\end{itemize}

Fig. \ref{fig:in-comp} shows the performance comparisons on the SBU, NTU (Cross-Subject), NTU (Cross-View) datasets respectively. We can see in comparison with the baseline scheme LSTM, the introduction of the spatial attention module (SA-LSTM) and the temporal attention module (TA-LSTM) brings up to 5.1\% and 6.4\% accuracy improvement, respectively. The best performance is achieved by combining both modules (STA-LSTM).

In the objective function as defined in (\ref{equ:lossfuntion}), the second and the third items for regularizations are designed for the spatial attention and temporal attention model, respectively. We can see they improve the performance of both spatial attention model and temporal attention model.
% by avoiding the ill-cases.
%both spatial and temporal attention help improve the performance of action recognition from skeleton data.
% for our spatial, temporal, and our final models respectively
%\begin{table}[htbp]
%	\fontsize{8pt}{9pt}\selectfont\centering
%	\begin{center}
%		\caption{Performance evluation of our spatial and temporal attention models on two datasets in accuracy (\%).} %%%%%%%%%%%%%
%		\label{table:In-comp}
%		\vspace{-2mm}
%		\begin{tabular}{c|c|c|c}
%			\hline
%			\multirow{2}{*} {Methods} & \multirow{2}{*}{SBU} & \multicolumn{2}{c}{NTU} \\
%			\cline{3-4} & & Cross Subject & Cross View \\
%			\hline
%			LSTM & 86.7 & 66.8 & 77.5\\
%			\hline
%			SA-LSTM & 88.0 & 71.9 & 80.4\\
%			\hline
%			TA-LSTM & 88.8 & 73.2 & 80.5\\
%			\hline
%			STA-LSTM & \textbf{91.5} & \textbf{73.4} & \textbf{81.2}\\
%			\hline
%			\hline
%			SA-LSTM (w/o regularization) & - & - & -\\
%			\hline
%			TA-LSTM (w/o regularization) & - & - & -\\
%			\hline
%		\end{tabular}
%	\end{center}
%\vspace{-5mm}
%\end{table}

\begin{comment}
\begin{table}[htbp]
\fontsize{8pt}{9pt}\selectfont\centering
\begin{center}
\caption{Accuracy(\%) comparisons for our spatial and temporal attention models on NTU dataset.} %%%%%%%%%%%%%
\label{table:In-comp}
\begin{tabular}{c|c|c|c}
\hline
Methods & SBU & Cross Subject) & Cross View \\
\hline
Baseline & 86.72 & 66.80 & 77.53\\
\hline
SA-LSTM & 88.00 & 71.89 & 80.35\\
\hline
TA-LSTM & 88.79 & 73.16 & 80.51\\
\hline
STA-LSTM & 91.51 & 73.42 & 81.23\\
\hline
\end{tabular}
\end{center}
\end{table}
\end{comment}

%\subsection{5.4~~Effectiveness of Regularization Terms}

\subsection{5.5~~Comparisons to Other State-of-the-Art}
%Evaluation of Our Proposed Final Scheme}

We show performance comparisons of our final scheme with the other state-of-the-art methods in Table \ref{table:SBU} and Table \ref{table:NTU} for the SBU and NTU datasets, respectively. Thanks to the introduction of the spatio-temporal attention models with efficient regularizations and the training strategy, our model is capable of extracting discriminative spatio-temporal features. We can see that our scheme achieves about 10\% accuracy gain on the NTU dataset for the Cross-Subject and Cross-View settings, respectively.
%\textcolor{red}{(4.2\% and 3.5\% gain ECCV2016)}
\begin{table}[htbp]
\vspace{-3mm}
	\fontsize{8pt}{9pt}\selectfont\centering
	\begin{center}
		\caption{Comparisons on the SBU dataset in accuracy (\%).} %%%%%%%%%%%%%
		\label{table:SBU}
		\vspace{-2mm}
		\begin{tabular}{c|c}
			\hline
			Methods & Acc. (\%) \\
			\hline
			Raw skeleton \cite{yun2012two} & 49.7 \\
			\hline
			Joint feature \cite{yun2012two} & 80.3 \\
			\hline
			Raw skeleton \cite{ji2014interactive} & 79.4 \\
			\hline
			Joint feature \cite{ji2014interactive} & 86.9 \\
			\hline
			Hierarchical RNN \cite{CVPR15HRNN} & 80.35 \\
			\hline
			Co-occurrence RNN \cite{zhu2015co} & 90.41 \\
			\hline
			%\hline
			%LSTM & 86.72 \\
			%\hline
			%SLSTM & 88.00 \\
			%\hline
			%TLSTM & 88.79 \\
			\hline
			STA-LSTM & \textbf{91.51} \\
			\hline
		\end{tabular}
	\end{center}
\vspace{-8mm}
\end{table}

\begin{table}[htbp]
	\fontsize{8pt}{9pt}\selectfont\centering
	\begin{center}
		\caption{Comparisons on the NTU dataset with Cross-Subject and Cross-View settings in accuracy (\%).} %%%%%%%%%%%%%
		\label{table:NTU}
		\vspace{-2mm}
		\begin{tabular}{c|c|c}
			\hline
			Methods & CS  & CV \\
			\hline
			Lie Group (Vemulapalli et al. 2014) & 50.1  & 52.8\\ %\cite{vemulapalli2014human}
			\hline
			Skeleton Quads (Evangelidis et al. 2014)  & 38.6 & 41.4\\ %\cite{evangelidis2014skeletal}
			\hline
			Dynamic Skeletons  \cite{hu2015jointly} & 60.2  & 65.2\\
			\hline
			HBRNN  \cite{CVPR15HRNN} & 59.1 &  64.0\\
			\hline
			%Deep RNN\cite{Shahroudy_2016_CVPR} & 56.3 & 64.1\\
			%\hline
			Deep LSTM \cite{Shahroudy_2016_CVPR} & 60.7  & 67.3\\
			\hline
			Part-aware LSTM   \cite{Shahroudy_2016_CVPR} & 62.9 & 70.3  \\
			\hline
			%\textcolor{red}{ST-LSTM (Tree) + Trust Gate \cite{liu2016spatio}} & 69.2  & 77.7\\ %ST-LSTM (Tree Traversal) + Trust Gate \cite{liu2016spatio}
			\hline
			%\hline
			%Baseline & 66.8  &  77.5 \\ %& 66.80  &  77.53
			%\hline
			STA-LSTM & \textbf{73.4} &  \textbf{81.2} \\ %73.42 &  81.23
			\hline
		\end{tabular}
	\end{center}
\vspace{-6mm}	
\end{table}

%\begin{table}[htbp]
%	\fontsize{9pt}{10pt}\selectfont\centering
%	\begin{center}
%		\caption{Comparisons on CMU dataset.} %%%%%%%%%%%%%
%		\label{table:CMU}
%		\begin{tabular}{c|c}
%			\hline
%			Methods & Acc. (\%) \\
%			\hline
%			Raw skeleton \cite{yun2012two} & 49.7 \\
%			\hline
%			Joint feature \cite{yun2012two} & 80.3 \\
%			\hline
%			Raw skeleton \cite{ji2014interactive} & 79.4 \\
%			\hline
%			Joint feature \cite{ji2014interactive} & 86.9 \\
%			\hline
%			Hierarchical RNN \cite{CVPR15HRNN} & 80.35 \\
%			\hline
%			Co-occurrence RNN \cite{zhu2015co} & 90.41 \\
%			\hline
%			\hline
%			Baseline & 00 \\
%			\hline
%			%SLSTM & 00 \\
%			%\hline
%			%TLSTM & 00 \\
%			%\hline
%			STLSTM & 00 \\
%			\hline
%		\end{tabular}
%	\end{center}
%\end{table}

%\vspace{-5mm}

\section{6.~~Conclusion}
We present an end-to-end spatio-temporal attention model for human action recognition from skeleton data. To select discriminative joints automatically and adaptively, we propose a spatial attention module with joint-selection gates to assign different importance to each joint. To automatically exploit the different levels of importance of different frames, we propose a temporal attention module to allocate different attention weights to each frame of the whole sequence. Finally, we design a joint training procedure to efficiently combine spatial and temporal attention with a regularized cross-entropy loss. Experimental results demonstrate the effectiveness of our proposed model which achieves remarkable performance improvement in comparison with other state-of-the-art methods.
%In this paper, we propose an end-to-end spatio-temporal attention based action recognition algorithm for skeleton data with the help of LSTM network. To select discriminative joints automatically, we conduct spatial attention mechanism to allocate different weights on each joint. Besides, temporal attention is utilized to indicate the importance on each frame of the whole sequence. Finally, the network is trained in an iterative manner to combine both spatial and temporal attention. Experimental results illustrates the effectiveness of our model by achieving remarkable performance compared with other state-of-the-art methods.

\bibliographystyle{aaai}
\small
\bibliography{egbib}
\begin{comment}
\begin{table}[htbp]
	\fontsize{8pt}{9pt}\selectfont\centering
	\begin{center}
		\caption{Performance evluation of our spatial and temporal attention models on two datasets in accuracy (\%).} %%%%%%%%%%%%%
		\label{table:In-comp}
		%\vspace{-2mm}
		\begin{tabular}{c|c|c|c}
			\hline
			\multirow{2}{*} {Methods} & \multirow{2}{*}{SBU} & \multicolumn{2}{c}{NTU} \\
			\cline{3-4} & & Cross Subject & Cross View \\
			\hline
			LSTM & 86.7 & 66.8 & 77.5\\
			\hline
			SA-LSTM & 88.0 & 71.9 & 80.4\\
			\hline
			TA-LSTM & 88.8 & 73.2 & 80.5\\
			\hline
			STA-LSTM & \textbf{91.5} & \textbf{73.4} & \textbf{81.2}\\
			\hline
			\hline
			SA-LSTM (w/o regularization) & - & - & -\\
			\hline
			TA-LSTM (w/o regularization) & - & - & -\\
			\hline
		\end{tabular}
	\end{center}
	%\vspace{-5mm}
\end{table}
\end{comment}
\end{document}